\documentclass[11pt,a4paper]{report}
\usepackage[dvipdfmx]{graphicx}
\usepackage[utf8]{inputenc}
\usepackage{color,graphicx}
\usepackage{amsmath,amssymb}
\usepackage{bm}
\usepackage{ascmac}
\usepackage{authblk}

\usepackage[lmargin=20mm,rmargin=20mm,tmargin=25mm, bmargin=20mm]{geometry}

\usepackage {booktabs}
\usepackage{balance} 
\usepackage{bm}
\usepackage{algorithmic}
\usepackage{algorithm}
\usepackage{multirow}

\newcommand{\EE}{\mathbb{E}}
\newcommand{\PP}{\mathbb{P}}

\newcommand{\HH}{\mathbb{H}}

\date{\today\\[30pt] Supervised by Prof. Takashi Goda}

\title{%
  \Large Bachelor thesis \\[30pt]
  \huge Sequential Bayesian experimental designs via reinforcement learning}

\author{Hikaru Asano}
\affil{Systems Innovation, Faculty of Engineering, The University of Tokyo }
\begin{document}

{
\makeatletter
\addtocounter{footnote}{1}
\renewcommand\thefootnote{\@fnsymbol\c@footnote}
\makeatother
\maketitle
}

\thispagestyle{empty}
\clearpage
\addtocounter{page}{-1}
\newpage


\begin{abstract}
Bayesian experimental design (BED) has been used as a method for conducting efficient experiments based on Bayesian inference. The existing methods, however, mostly focus on maximizing the expected information gain (EIG); the cost of experiments and sample efficiency are often not taken into account. In order to address this issue and enhance practical applicability of BED, we provide a new approach \emph{Sequential Experimental Design via Reinforcement Learning} to construct BED in a sequential manner by applying reinforcement learning in this paper. Here, reinforcement learning is a branch of machine learning in which an agent learns a policy to maximize its reward by interacting with the environment. The characteristics of interacting with the environment are similar to the sequential experiment, and reinforcement learning is indeed a method that excels at sequential decision making. 

By proposing a new real-world-oriented experimental environment, our approach aims to maximize the EIG while keeping the cost of experiments and sample efficiency in mind simultaneously. We conduct numerical experiments for three different examples. It is confirmed that our method outperforms the existing methods in various indices such as the EIG and sampling efficiency, indicating that our proposed method and experimental environment can make a significant contribution to application of BED to the real world.
\end{abstract}

\tableofcontents
\clearpage\setcounter{page}{0}\pagenumbering{roman}\pagestyle{plain}
 \newpage

\setcounter{page}{0}\pagenumbering{arabic}
\chapter{Introduction}
Experimentation is one of the essential elements of the scientific method. We collect data by conducting experiments, and we discover scientific knowledge based on the data. Experiments are very practical for science and engineering, but they are also often very costly. Regardless of the discipline such as physics, psychology, and sociology, conducting experiments can be expensive in terms of both cost and time. For example, a large-scale physics experiment requires a lot of money to conduct a single experiment. Also, drug trial involves the use of an uncertain drug, so there is a need to obtain a lot of knowledge while minimizing the cost in terms of side effects to the subjects. As a result, we often want to discover as much as possible with as little cost. In other words, there are efficient experiments and inefficient experiments, and we want to design and conduct efficient experiments. The efficiency of such an experiment can be viewed from the perspective of how to minimize the cost of the experiment, and also from the perspective of how to maximize the information obtained within a limited cost.  For example, in human experiments, the number of data may be small due to the difficulty of collecting subjects.In the case of a human experiment, for example, the number of data can be reduced due to the difficulty of collecting subjects. In a situation where data is difficult to obtain, designing an efficient experiment is also an important aspect of the efficiency of the experiment.

The design of efficient experiments has been studied in the field of experimental designs \cite{Fisher1966,Lindley1956OnAM,Chaloner1995BayesianED,Myung2013ATO,santner2003design}.
In the past, methods such as those using Fisher information matrix were proposed for constructing good experimental designs; there are many alphabetic optimality criteria known in the literature, see \cite{10.5555/1206386}. However, these methods are not necessarily effective for non-linear models \cite{10.1214/aos/1032526965,ryan2016review}.
In recent years, in order to design efficient experiments for nonlinear models, much attention has also been focused on methods based on Bayesian inference. Using Bayesian inference, it is possible to update the probability distribution of model's parameters conditional on the data gathered from an experiment and to evaluate the goodness of the experiment by using the mutual information between the model before and after the update, which measures how much the uncertainty of parameters can be reduced on average by the experiment. The amount of the mutual information between the pre-and post-experiment models is generally referred to as the expected information gain (EIG). The experimental design maximizing the EIG is called (optimal) Bayesian experimental design (BED) \cite{Lindley1956OnAM,Chaloner1995BayesianED,Ryan2003EstimatingEI,Ryan2014TowardsBE,Ryan2015FullyBE,Shababo2013BayesianIA,Vanlier2012ABA}. BED has been mainly used in the batch form, in which all the individual experiments are determined in advance \cite{HUAN2013288,doi:10.1080/01621459.1995.10476636}. However, it is more natural to think that efficient experimental conditions are not entirely known in advance but become apparent gradually. 

Sequential design, in which experimental conditions are determined sequentially, is attracting attention to solve the problem of the batch design. The key difference from the batch design is that we optimize a policy (instead of the design itself) on how to set the experimental condition depending on the earlier conditions and the corresponding observations in an iterative manner. Sequential design often lead to experiments more efficiently than the batch design in environments where data is available sequentially \cite{foster2021deep,huan2016sequential,Kleinegesse2020SequentialBE}. However, the sequential design also has problems. The existing methods that construct sequential designs may not be easy to apply to complex environments because the experimental environment is often restricted to simple model settings \cite{McGree2012AdaptiveBC,10.2307/2533849,Pronzato2002,10.2307/2291016}. In addition, most of the existing methods require to compute the EIG during each sequential step based on the data observed so far. In general, the cost of such computation is very high \cite{Foster2019VariationalBO,Goda2018MultilevelMC}, which makes it challenging to 
apply in situations where real-time experiment planning is required. To solve this problem, a method of planning experiments according to previously learned policies is proposed \cite{foster2021deep,huan2016sequential}.

The conventional methods to construct sequential experimental designs are efficient in maximizing the EIG, but may not be good at reducing the cost of experiments. For example, consider an experiment to find a sound source by sampling sound intensity pointwise while moving on the ground. In such an experiment, the location of the sound source is expected to be most accurately identified by optimizing the sequence of sample points so that the EIG is maximized. To make the cost of experiment small, on the other hand, the sequence of sample points should be designed in a way that the distance traveled is minimized.
Minimizing the cost is also important in the following aspects: it can increase the number of experimental trials to obtain more accurate estimates, and can deepen the analysis by conducting experiments in different environments. 

In addition, it is desirable to make the amount of data required to learn a policy for designing experiments small \cite{munos2016safe,Wang2017SampleEA}. Any method requiring a large amount of training data cannot be applied to practical complex experimental environments.
This way, to apply experimental designs in real experiments, it is important to consider the cost of the experiment and the efficiency of policy learning simultaneously, as well as maximizing the EIG.
In this study, by applying reinforcement learning and proposing a novel real-world-oriented experimental environment, we provide a new approach to construct sequential BED achieving these goals at the same time, which can enhance practical applicability of BED substantially.

\chapter{Related works}
The experimental designs have been studied in terms of how to allocate resources for experiments, and classical methods have been proposed to construct optimal designs using the Fisher information matrix \cite{pukelsheim1980,AtkinsonDonev1992}. Such methods are quite effective for linear models but can only lead to local optimal solutions for non-linear models \cite{10.1214/aos/1032526965,ryan2016review}.
On the contrary, the BED represents the amount of information obtained from an experiment by using Bayesian updating. More precisely, the BED employs the EIG as a quality criterion of the experimental design, which coincides with the (expected) mutual information between the prior and posterior probability distributions of model parameters \cite{Lindley1956OnAM,Chaloner1995BayesianED}. This method is also effective for non-linear models and has been applied to various problems \cite{Ryan2003EstimatingEI,10.2307/2533849,Cook2008OptimalOT,Ryan2014TowardsBE,Ryan2015FullyBE,Shababo2013BayesianIA,Vanlier2012ABA}. 
The major technical difficulty of the BED is in that computing the EIG is not easy when Bayesian updates of probability distributions cannot be given in a closed-form. To approximate the EIG or the poster distribution, one can use Markov chain Monte Carlo method \cite{Amzal2006BayesianOptimalDV}, importance sampling including Laplace approximation \cite{Long2013FastEO,Beck2018FastBE,Beck2018MultilevelDL}, variational inference \cite{Foster2019VariationalBO} and multilevel Monte Carlo method \cite{giles2008multilevel,Goda2018MultilevelMC,Beck2018MultilevelDL}. However, the cost of performing these computations is still not sufficiently low for practical applications.

The experimental designs can be categorized into batch designs and sequential designs. 
The batch designs are effective when the experimental data are obtained in a batch manner \cite{HUAN2013288,doi:10.1080/01621459.1995.10476636}.  However, in many experiments, the observed data is obtained sequentially and the sequential designs naturally fit better than the batch designs \cite{huan2016sequential,Kleinegesse2020SequentialBE,foster2021deep}. 
However, the existing works on sequential designs have been hampered by the myopic nature of the policy \cite{McGree2012AdaptiveBC}, the low degree of freedom in policy \cite{Rossell2013}, and the computational cost of the policy \cite{huan2016sequential}. Moreover, the existing construction methods require to compute the EIG in every sequential step \cite{pmlr-v119-jiang20b,Kleinegesse2020SequentialBE}, which makes it hard to apply for real-time sequential design. Recently a sophisticated method has been proposed to enable real-time decision making by learning the policy in advance for designing experiments \cite{foster2021deep}. Our current work builds on it, and applies the framework of reinforcement learning to deal with the cost of experiment and sample efficiency simultaneously.
\chapter{Formulation}
In this section, we give overviews on the BED and reinforcement learning, respectively.

\section{Bayesian Experimental Design}
Recently the BED has been studied quite intensively as a way of experimental designs to tackle with non-linear complex models. In what follows, we first formulate the EIG as a quality criterion of the BED, and then give its contrastive bounds. Moreover, we extend the EIG to the sequential data generating process.

\subsection{Expected Information Gain}
The goal of the BED is to find the optimal design (for the batch case) or policy (for the sequential case) $\pi =\pi^{*}$ that minimizes the information entropy of the latent variable $\theta$ in a model by conducting experiments based on the condition $\pi$. This corresponds to maximizing the EIG whose formulation for the batch design is given as follows.

Before conducting the experiment, the information entropy of the latent variable, $H[p(\theta)]$, is given by
\begin{align} \label{theta_pri_ent}
  H[p(\theta)]&= - \int_{\theta} p(\theta) \log p(\theta) d\theta \\
  &=-\EE_{\theta}[\log p(\theta)]], 
\end{align}
where $p(\theta)$ denotes the prior probability distribution of the latent variable $\theta$. The following Bayes formula expresses that the distribution of $\theta$ is updated by conducting an experiment based on the design $\pi$ and obtaining the observation $y$:
\begin{align} \label{theta_post_dist}
  p(\theta | \pi,y)= \frac{p(y|\theta,\pi)p(\theta)} {p(y|\pi)}.
\end{align}
Here, $p(y|\theta,\pi)$ describes a generative model of the observable $y$ given $\theta$ and $\pi$. It follows from \eqref{theta_post_dist} that the information entropy of the posterior distribution, $H\left[p(\theta|\pi,y)\right]$, is given by
\begin{align} \label{theta_post_ent}
  H[p(\theta|\pi,y)] &= \int_{\theta} p(\theta|\pi,y) \log p(\theta|\pi,y) d\theta \notag \\
  &= \int_{\theta} p(\theta|\pi,y) \log \frac{p(y|\theta,\pi)p(\theta)} {p(y|\pi)} d\theta \notag \\
  &=-\EE_{p(\theta|\pi,y)}\left[\log\frac{p(y|\theta,\pi)p(\theta)} {p(y|\pi)}\right] .
\end{align}

The difference between \eqref{theta_pri_ent} and the expectation of \eqref{theta_post_ent} with respect to $y$ is regarded as the amount of information gained by the BED $\pi$, and is nothing but the EIG:
\begin{align} 
 U(\pi) &:=\EE_{p(y|\pi)} \left[H[p(\theta)]-H[p(\theta|\pi,y)]\right] \notag \\ 
 &= -\int_y \int_{\theta} p(y|\pi)  p(\theta) \log p(\theta) d\theta dy + \int_y \int_{\theta} p(y|\pi)  p(\theta|\pi,y) \log \frac{p(y|\theta,\pi)p(\theta)} {p(y|\pi)} d\theta  dy \notag \\
 &= -\int_y p(y|\pi) dy \int_{\theta} p(\theta) \log p(\theta) d\theta + \int_y \int_{\theta} p(y|\pi) \frac{p(y|\theta,\pi)p(\theta)} {p(y|\pi)} \log \frac{p(y|\theta,\pi)p(\theta)} {p(y|\pi)} d\theta dy \notag \\
 &=- \int_{\theta} p(\theta) \log p(\theta) d\theta + \int_y \int_{\theta} p(y|\theta,\pi)p(\theta)  \log \frac{p(y|\theta,\pi)p(\theta)} {p(y|\pi)} d\theta dy \notag \\
 &= - \int_y\int_{\theta}  p(y|\theta,\pi)p(\theta) \log p(\theta) d\theta dy + \int_y \int_{\theta} p(y|\theta,\pi)p(\theta)  \log \frac{p(y|\theta,\pi)p(\theta)} {p(y|\pi)} d\theta dy \notag \\
 &= \EE_{p(y| \theta,\pi)}\EE_{p(\theta)} \left[ -\log p(\theta) + \log \frac{p(y|\theta,\pi)p(\theta)}{p(y|\pi)} \right] \notag \\
 & \: = \EE_{p(y| \theta,\pi)}\EE_{p(\theta)} \left[ \log p(y | \theta,\pi)-\log p(y|\pi) \right].  \label{eig}
\end{align}

\subsection{Contrastive Information Bounds}
As already stated, the goal of the BED is to find a design $\pi=\pi^{*}$ that maximizes the EIG $U(\pi)$. However, calculating \eqref{eig} is generally challenging. This is because, in order to evaluate $p(y|\pi)$ exactly, it is necessary to perform the integral calculation $\int_{\theta} p(y|\pi,\theta)p(\theta)d\theta$. These integrals are in general difficult to compute analytically. In such cases, it is necessary to obtain the EIG by approximation. In order to obtain an unbiased estimate of the EIG, it is necessary to take the nested expectation, as is clear from \eqref{eig}. 
And finding the nested expectation value is generally known to be computationally expensive. 

For example, if we want the root mean square error to be $\epsilon$ for the EIG using the Monte Carlo method, we need $O(\epsilon^{-3})$\cite{Rainforth2018OnNM,Beck2018FastBE,Ryan2014TowardsBE}. Methods using multilevel Monte Carlo\cite{giles2008multilevel,giles2013multilevel,Giles2015MultilevelMC,Beck2018FastBE,Goda2018MultilevelMC,Rainforth2018OnNM,Beck2018MultilevelDL} or a combination of Laplace approximation and importance sampling\cite{Long2013FastEO} have been proposed to perform such calculations efficiently.  However, even using such a method, the computational complexity of $O(\epsilon^{-2}(\log \epsilon^{-1})^2)$ or $O(\epsilon^{-2}(\log \epsilon^{-1})^2)$ is required to compute the nested expectation calculation. The method to approximate the EIG using variational inference has also been proposed\cite{Foster2019VariationalBO}. However, this method is also computationally expensive because it requires optimization for approximation.

In general, finding the optimal design $\pi^{*}$ requires iterative optimization. The large computational cost of the pointwise EIG evaluation means that the overall computational cost of optimization is enormous \cite{DBLP:conf/icml/ZhengPF18,Rainforth2018OnNM}. To address this issue, an unbiased estimator for the gradient $\nabla_{\theta}U(\theta)$ has been introduced and combined with stochastic gradient-based optimization algorithms to search for $\pi^*$ efficiently in \cite{DBLP:journals/corr/abs-2005-08414}. However, the research outcomes of this direction are still immature, so that we employ a computationally-cheap, lower bound on the EIG as a quality criterion in this paper.


To be precise, we use contrastive information bounds (CID), which have proven quite effective in a recent work \cite{foster2021deep}. Here CID can be computed as follows. First, we sample the latent variable $\theta_0 \sim p(\theta)$ and generate the observation $y$ by following the generating process determined by $p(y\mid \theta_0,\pi)$.
Then, we generate $L$ independent contrastive samples $\theta_{1:L} \sim p(\theta)$. By using the initially sampled $\theta_0$, the observation $y$, and the $L$ contrastive samples $\theta_{1:L}$, CID is given by
\begin{align}
  \mathcal{I}_L=\log \frac{p(y|\theta_0,\pi)}{\frac{1}{L+1}\sum^L_{l=0}p(y|\theta_l,\pi)}.
\end{align}
The expected value of CID, $\mathcal{L}$, is defined by
\begin{align}\label{ecid}
  \mathcal{L}(\pi,L)=\EE_{p(\theta_{0:L})}\EE_{p(y |\theta_0,\pi)}\left[ \log \frac{p(y|\theta_0,\pi)}{\frac{1}{L+1}\sum^L_{l=0}p(y|\theta_l,\pi)}\right].
\end{align}

\subsection{Extended Expected Information Gain}
So far, we have formulated the EIG for the batch design $\pi$. We can easily modify the formulation for the sequential design in which $\pi$ denotes a policy and the observation $y$ is generated sequentially. First, let us rewrite the EIG given in \eqref{eig} into the plural form:
\begin{align}
U(\pi)=\EE_{p(y_{0:t}| \pi)} \EE_{p(\theta | y_{0:t},\pi)}\left[ \log p(y_{0:t} | \theta,\pi)-\log p(y_{0:t}|\pi)  \right] \label{seig}  
\end{align}
where $y_{0:t}$ represents a set of the observations given discretely from time $0$ to time $t$. 

Similarly, the CID can be modified into the plural form. As before, we first sample the latent variable $\theta_0 \sim p(\theta)$. Then, according to the policy $\pi$, we obtain the initial experimental design $\xi_0$ and generate the corresponding observation $y_0$. Depending on this result, we obtain the second experimental design $\xi_1$ and generate the corresponding observation $y_1$. Repeating this procedure, we can get a sequence of the experimental designs $\xi_{0:t}$ and the observations $y_{0:t}$. Now the CID for the sequential case is given by
\begin{align}\label{scid}
  \mathcal{I}_L=\log \frac{p(y_{0:t}|\theta_0,\xi_{0:t})}{\frac{1}{L+1} \sum^L_{l=0}p(y_{0:t}|\theta_l,\xi_{0:t})}.
\end{align}
Note that $p(y_{0:t}|\theta,\xi_{0:t})$ in \eqref{scid} is conditionally independent if the observable $y_{t}$ depends only on the corresponding design $\xi_{t}$. In such cases, $p(y_{0:t}|\theta,\xi_{0:t})$ can be expressed as
\begin{align}
 p(y_{0:t}|\theta,\xi_{0:t})= \prod^{t}_{i=0}p(y_{i}|\theta,\xi_{i}).
\end{align}


\section{Reinforcement Learning}
Reinforcement learning is a research field that aims to find an optimal decision rule (policy) for sequential decision making. It is also a branch of machine learning characterized by the notion of reward $r$, where decision rules are learned to maximize the expected value of the reward. The two most essential elements in reinforcement learning are the agent and the environment. The agent tries to maximize the reward given to itself by interacting with the environment. 
The agent is first given an initial state $s_0\in{\mathcal{S}}$ by the environment to interact with. Given a state from the environment, the agent outputs an action $a_t\in{\mathcal{A}}$ based on its own policy at each time $t$ according to $a_t \sim \pi(a|s_t)$. The environment then returns the next state $s_{t+1}$ and the reward $r_t \in{\mathcal{R}}$ to the agent based on the transition function $\PP(s_{t+1},r_t | s_t,a_t)$. 
Through these interactions, the agent learns the policy $\pi^{*}$ that maximizes the expected reward $V^{\pi}(s)$ defined by
\begin{align}\label{value}
  V^{\pi}(s)=\EE \left[ \sum^{n}_{k=0} \gamma^{k}r(s_{t+k},a_{t+k}) \mid s_t=s \right],
\end{align}
where $\gamma$ represents the decay rate. \eqref{value} is called the value function. 

It is also possible to define Q-value function by adding the action $a_t$ at time $t$ to the conditions of the value function as follows:
\begin{align}\label{Qfunc}
  Q^{\pi}(s,a)=\EE \left[ \sum^{n}_{k=0} \gamma^{k}r(s_t,a_t)  \mid s_t=s,a_t=a \right].
\end{align}
Here the evaluation for a state/action pair requires $n+1$ steps of information. The number of states that need to be considered to find the optimal policy grows exponentially with $n$. Performing such calculation is impractical. In reinforcement learning, the famous Bellman equation is used to compute \eqref{Qfunc} efficiently. Using the Bellman equation, \eqref{Qfunc} can be expressed as follows.
\begin{align}
 Q^{\pi}(s_t,a_t)=r(s_t,a_t)+\gamma \EE\left[ V^{\pi}(s_{t+1}) \right].
\end{align}
By using the Bellman equation, the information required to compute the Q-value is significantly reduced, that is, $s_t,a_t$ and $s_{t+1}$.

Reinforcement learning can be divided into two main approaches in how it learns the policy: value-based approach and policy-based approach. In the value-based approach, the agent approximates the Q-value in order to take the optimal action, and takes action based on this function. For example, the greedy agent determines the action as $a_t=\mbox{argmax}_{a\in{\mathcal{A}}}Q^{\pi}(s_t,a)$. A typical example of the value-based approach is DQN (Deep Q Network) \cite{Mnih2015HumanlevelCT,10.5555/3016100.3016191,DBLP:journals/corr/SchaulQAS15,pmlr-v48-wangf16,fortunato2018noisy,Bellemare:2017:DPR:3305381.3305428,Hessel2018RainbowCI}, which has been applied to various environments such as video games with high results. The value-based approach is known to be effective when the action space $\mathcal{A}$ takes discrete values. However, its performance is known to be degraded when the action space $\mathcal{A}$ takes continuous values, such as in robot control \cite{Duan2016BenchmarkingDR}.

On the other hand, the agent learns its policy directly in the policy-based approach. A typical example of this approach is REINFORCE \cite{DBLP:journals/ml/Williams92}, which trains a policy network by approximating the Q function by taking the expected values of rewards obtained from multiple episodes. Although REINFORCE is a straightforward idea to learn a policy network using actual observed rewards, the training becomes unstable due to the variability in the rewards used to train the policy network. In order to solve this problem, the Actor-Critic algorithm \cite{Konda1999ActorCriticA,Amzal2006BayesianOptimalDV} uses two networks, the critic-network to evaluate the policy and the actor-network to decide the agent's action. This way the instability of policy evaluation occurring in REINFORCE is mitigated, and the learning process can be stabilized. Furthermore, reinforcement learning has a problem in the learning process that it becomes difficult to improve the reward once the policy is degraded. Some methods have been proposed to stabilize the learning process by limiting the amount of parameter updates during the update process \cite{Schulman2017ProximalPO,pmlr-v37-schulman15}. 

In the policy-based approach, a policy gradient is often used to compute the gradient $g$ for the policy network as follows \cite{Schulman2016HighDimensionalCC}:
\begin{align}
  g = \EE \left[ \sum^{\infty}_{t=0}\Phi(s_t,a_t) \nabla_\phi \log \pi_{\phi}(a_t|s_t) \right],
\end{align}
where $\phi$ represents the parameters of the policy network, and $\Phi(s_t,a_t)$ represents the value of the state at time $t$. We can use various functions for $\Phi(s_t,a_t)$, but the advantage function is often used to reduce the variance. As shown below, we can define the advantage function using the value function and the Q-value function
\begin{align}
  A^{\pi}(s_t,a_t)=Q^{\pi}(s_t,a_t)-V^{\pi}(s_t).
\end{align}

When training the agent network, many policy gradient algorithms use the method called on-policy learning, where the policy for determining the agent's action and the policy for updating the network need to be the same \cite{mnih2016asynchronous,Schulman2017ProximalPO,pmlr-v37-schulman15}. When using such a method, once the network is trained, the data used for training cannot be reused. In reinforcement learning, sample efficiency refers to how efficiently we can use samples in such training. The sample efficiency of on-policy learning is poor because the samples used for training needs to be discarded after each training step. In recent years, off-policy learning has been proposed for policy gradients, but these methods are not practical because the learning results strongly depend on the hyperparameters \cite{henderson2018deep}. One of the reasons why the learning results depend on hyperparameters is the low search capability. Due to it, the agent converges to a local optimum solution that depends on the hyperparameters. Low search capability also leads to poor performance in high-dimensional environment.

The Soft Actor-Critic (SAC) method was proposed as a solution to these problems \cite{pmlr-v80-haarnoja18b}. To improve the search capability, the entropy term of the policy is added to the objective function in SAC, as described in detail later.

In this paper, we use reinforcement learning as a method for sequential experimental design because it is desirable to apply experimental designs to more complex and large-scale experimental environments where the action space takes continuous values. In addition, it is necessary to assume that the computational cost of simulating the experiment is exceptionally high. We propose an experimental design based on SAC as a method that satisfies these requirements.


\chapter{Experimental Design via Reinforcement Learning}
In this section, we explain some concepts to explain the problems of the existing approaches. Then, our proposed method \emph{Sequential Experimental Design via Reinforcement Learning} is described.

\section{Batch Design and Sequential Design}
We have described the batch design as a method of determining all the experimental designs in advance, whereas the sequential design as a method of planning an experiment sequentially by a policy based on the data obtained from the earlier experiments. Therefore, the differences of batch and sequential designs can be found in the conditions given to $\pi$.
Now, consider the case where $T$ experiments are conducted between time 0 and time $T-1$. In such a case, in batch design, $\pi$ coincides with a set of the individual designs $\xi_{0:T-1}$ at the times $t=0,1,\ldots,T-1$.
In contrast, in sequential design, the design $\xi_{t}$ at each time $t$ is sampled according to the policy $\pi$ by
\begin{align}
  \xi_t \sim \pi(\xi |\theta,s_{0:t-1}),
\end{align}
where $s_{t}$ denotes the pair $(\xi_t,y_t)$ of the design and the observation at each time $t$. Since the experiment at $t$ is designed based on the state $s_{0:t-1}$ in sequential design, it is more flexible than batch design. 

\section{Myopic and Holistic Views}
Although sequential design is often more effective than batch design, it is challenging to make holistic decisions in sequential design. In other words, at each time $t$, if the design of the experiment $\xi_t$ prioritizes the reward $r_t$, i.e., the EIG, at that time only, the reward obtained in the whole experiments might decrease. Therefore, we need to formulate the difference between myopic and holistic experimental designs by using the EIG.

First, a myopic experimental design can be thought of as a policy that maximizes only the difference between the information entropy of the current $\theta$ and that of $\theta$ when a new state $s_t$ is obtained. Formally, the myopic experimental design is equal to a policy that maximizes the following objective function $U_t$ at each time $t$.
\begin{align}\label{myopic}
  U_t=H \left[P(\theta|s_{0:t-1}) \right]-H \left[P(\theta|s_{0:t}) \right]
\end{align}
An experimental design that only considers the amount of information we can obtain during one step can be regarded as a myopic design. Designs that are optimized by an objective function such as \eqref{myopic} are likely to converge to a locally optimal solution. In fact, the previous work shows that the designs trained by myopic rewards, such as the above, perform worse than the designs trained by holistic rewards \cite{huan2016sequential}. In addition, when considering rewards such as \eqref{myopic}, Bayesian updating must be performed at each time $t$, which is generally very computationally expensive.

To solve the problems of greediness in experimental design and reward calculation, the preceding study showed that the amount of information obtained between each step of an experiment sums up to the amount of information obtained in the entire experiment, and further formulated the information obtained in the whole experiment \cite{foster2021deep}. This means that there is the following relationship between the sum of information obtained between each step of the experiment and the amount of information obtained in the entire experiment:
\begin{align}
\EE \left[ \sum^{T-1}_{t=0}U_t \right]= H \left[P(\theta) \right]-H \left[P(\theta|s_{0:T-1}) \right].
\end{align}
In addition, by developing the information obtained in the whole experiment in a differentiable form, a novel algorithm called \emph{deep adaptive design} (DAD) has been proposed \cite{foster2021deep}. 

By using a contrastive lower bound on the amount of information obtained in the entire experiment as given in \eqref{ecid}, the training of DAD is performed by the reparametrized gradient
\begin{align}\label{dad}
  \frac{d \mathcal{L}_{T}\left(\pi_{\phi}, L\right)}{d \phi}=\mathbb{E}_{p\left(\theta_{0: L}\right) p\left(\epsilon_{0: T-1}\right)}\left [\frac{d I_{L}\left(\theta_{0: L}, s_{0:T-1}\right)}{d \phi}\right]
\end{align}
where $\epsilon_{0:T-1}$ is the observation error in the observation of the experiment.

\section{Cost of Experiment}
The equation \eqref{dad} allows us to formulate a sequential experimental design from a holistic perspective. Since the objective function \eqref{dad} is permutation-invariant with respect to the state $s_{0:T-1}$, DAD exploits this property for efficient and effective training \cite{foster2021deep}. For example, suppose that the experimental designs $\xi_s, \xi_t$, and $\xi_u$ are performed at times $s, t$, and $u$, respectively. The value of the objective \eqref{dad} does not change even if the order in which the experiment is performed is not the same as $s, t$, and $u$. 

In order to take the cost of experiments into account more carefully, however, we do not exploit this permutation-invariance property of the objective deliberately in this study. For example, when collecting three experimental data by moving a sampling point on the ground, it is desirable to sample the data in order of distance from the starting point to reduce the total cost of  experiment. This aspect is not respected if only a permutation-invariant objective is used.

To consider the sequential nature of the design of  experiment, it is desirable to express the reward of the experiment in the form of dynamic programming as formulated in \cite{huan2016sequential}. By setting up such an objective function, we can acquire a holistic perspective and, at the same time, consider the order in which the experiments are conducted. However, in the previous work \cite{huan2016sequential}, the state space $\mathcal{S}$ is delimited into a grid to calculate the objective function based on dynamic programming, and also the objective function is calculated using the form of Riemann integration. Such a method should be only useful when the state space $\mathcal{S}$ is low-dimensional or takes discrete space. To solve these problems, methods to approximate the objective function using neural networks or other methods have been used in reinforcement learning, and this paper also follows this approach.

\section{Sequential Experimental Designs via Reinforcement Learning}
In reinforcement learning, an agent learns a reward-maximizing policy $\pi$ by interacting with its environment. We design experiments by interacting with the experimental environment. The characteristics of this type of experimental design are very similar to the agent-environment relationship in reinforcement learning. Based on the above features, we consider the experiment as an environment in reinforcement learning.

To treat the experimental design as the environment in reinforcement learning, we need to introduce several definitions as below. First, the action $a_t$ in the experimental design is thought of as the amount of change between consecutive designs $\xi_{t-1}$ and $\xi_t$. In this setting, the current design $\xi_t$ is given by
\begin{align}\label{eq:design_diff}
\xi_t=\xi_{t-1}+a_t.
\end{align}
For example, if we take samples by moving an observation point in space, the action corresponds to the amount of movement from the current point, and the action space $\mathcal{A}$ can be given by (a subspace of) the three-dimensional Euclidean space. Then, the state $s$ in the experimental design corresponds to the pair of the design and the observation result. So, the current state $s_t$ is given by $s_t=(\xi_t,y_t)$, and the observed value $y$ is by $y=p(y|\xi,\theta)$.

Finally, we define the reward function. In this work, the amount of information obtained by an experiment is measured by using CID \eqref{scid} from the perspective of computational cost. Nevertheless, computing CID during each time step is computationally expensive in learning process. To reduce such costs, we calculate the CID only at the end of the experiment, i.e., at the end of an episode. Therefore, the reward is computed as follows:
\begin{align}
   r(s_{0:T},a_{0:T})=\left\{\begin{array}{ll} 0 &\mbox{if episode is not finished,} \\
\mathcal{I}_L(s_{0:T},a_{0:T}) & \mbox{if episode is finished.} \end{array} \right. 
\end{align}
Note that calculating the CID only at the end of the experiment and using it as the objective function is the same as in DAD. 
To design a sequential experiment that simultaneously takes both the EIG and the cost of the experiment, it is necessary to set up a sequential objective function. We do this by applying reinforcement learning in this study. 

Additionally, this paper aims to provide effective real-time experiment plannings for complex and larger-scale problems. To achieve this goal, SAC is considered to be effective because of its high sample efficiency and high performance, even for continuous-valued action spaces. SAC can reduce the impact of hyperparameters on learning by increasing the search capability. The basic idea to enhance the search capability is based on soft Q-learning \cite{pmlr-v70-haarnoja17a}, which adds an entropy term to the usual Q-value function \eqref{Qfunc}. The objective function in soft Q-learning is given by
\begin{align}\label{soft-q}
   Q^{\pi}(s,a)=\EE \left[ r_t+\sum^{T}_{k=1} \gamma^{k}(r_{t+k}+\alpha \HH(\pi(\cdot|s_{t+k})) \mid s_t=s, a_t=a,\pi \right],
\end{align}
where $\HH(\pi(\cdot|s_{t+k}))$ denotes the entropy of a policy. Maximizing the entropy of a policy means that it is more likely to take a different action $a_t$ given the same state $s_t$. Soft Q-learning maximizes the reward and the entropy at the same time, as in \eqref{soft-q}, which makes it possible to take a balance between exploration and exploitation. In addition, the parameter $\alpha$ in \eqref{soft-q} represents the entropy temperature, which adjusts the balance between search and exploitation during the training. The Bellman equation can be adapted to \eqref{soft-q} as well as in standard Q-learning.

\begin{align}
 Q^{\pi}(s_t,a_t) & = r(s_t,a_t)+\gamma \EE\left[ V(s_t) \right], \\
 V(s_t) & = \EE_{a_t \sim \pi}\left[Q(s_t,a_t)-\alpha \log\pi(a_t,s_t)\right].
\end{align}
As you can see, the objective function in soft Q-learning can be expressed by adding an entropy term to the usual value function.

SAC is a method that applies the soft Q-learning to the actor-critic algorithm.  As in the usual actor-critic algorithm, SAC is composed of two networks: the actor-network and the critic-network.
The critic-network is trained by using the squared residual error
\begin{align}
  J_Q(\psi)=\EE\left[\frac{1}{2}\left(Q_{\psi}\left(s_{t}, a_{t}\right)-\left(r\left(s_{t}, a_{t}\right)+\gamma E_{s_{t+1} \sim p}\left[V_{\bar{\psi}}\left(s_{t+1}\right)\right]\right)\right)^{2}\right],
\end{align}
where $\psi$ denotes the set of the parameters of the critic-network and $\bar{\psi}$ denotes that of the target network. The target network is a network that is used only when training the critic-network. It is known that the use of the target network stabilizes the training process \cite{10.5555/3016100.3016191}.

The actor-network is trained by minimizing the KL-divergence with the softmax of \eqref{soft-q}. Therefore, the objective function of policy$\pi$ can be expressed as follows:
\begin{align}
  J_{\psi}(\phi)=\EE_{a_{t} \sim \pi}\left[D_{K L}\left(\pi_{\phi}\left(\left.\cdot\right|_{S_{t}}\right) \| \frac{\exp \left(\frac{1}{\alpha} Q_{\psi}\left(s_{t}, \cdot\right)\right)}{Z_{\psi}\left(s_{t}\right)}\right)\right],
\end{align}
where $\phi$ denotes the set of the parameters of the actor-network and $Z_{\psi}\left(s_{t}\right)$ denotes the normalization constant of the softmax of \eqref{soft-q}. Here, $ Z_{\psi}\left(s_{t}\right)$ can be ignored since it is independent of $\phi$. Therefore, the objective function of the actor-network can be expressed more simply by
\begin{align}
  J_{\psi}(\phi)=\EE_{a_{t} \sim \pi}\left[\log \pi_{\phi}\left(a_{t} \mid s_{t}\right)-Q_{\psi}\left(s_{t}, a_{t}\right)\right].
\end{align}

The whole process of our algorithm is described in Algorithm \ref{alg1}.

\begin{figure}[ht]
\begin{algorithm}[H]
    \caption{\emph{Sequential Experimental Design via Reinforcement Learning}}
    \label{alg1}
    \begin{algorithmic}   
    \STATE Initialize policy and value network parameters $\phi,\psi$
    \STATE Initialize target network parameters $\bar{\psi}$
    \STATE Set learning rates $\lambda_Q, \lambda_{\pi}$ and weight $\tau$
    \FOR{iteration = 1,2,...}
        \STATE sample latent variable $\theta_0$
            \WHILE{episode is finished}
                \STATE Select action $a_t \sim \pi(a_t|s_t)$ and receive reward $r_t$ and next state$s_{t+1}$
                \STATE Store transition$(s_t,a_t,r_t,s_{t+1})$ in $R$
                \STATE sample $(s_t,a_t,s_{t+1},r_t)$ from replay buffer$R$
                \STATE $\psi \leftarrow \psi-\lambda_{Q} \nabla_{\psi} J_{V}(\psi)$
                \STATE $\phi \leftarrow \phi-\lambda_{\pi} \nabla_{\phi} J_{V}(\phi)$
                \STATE $\bar{\psi} \leftarrow \tau \psi+(1-\tau) \bar{\psi})$
            \ENDWHILE
    \ENDFOR
    \end{algorithmic}
\end{algorithm}
\end{figure}

\chapter{Experiment}
The main goal of this paper is to bring experimental designs closer to real-world applications. In order to achieve this goal, it is desirable to construct experimental designs not only sequentially but also in real-time. In the earlier works, many sequential construction algorithms have been proposed \cite{pmlr-v119-jiang20b,Kleinegesse2020SequentialBE}, but most of them require a large computational cost to make designs and are not necessarily suitable for real-time decision making. DAD proposed in \cite{foster2021deep} is a good algorithm for policy optimization because it is built upon an computationally-cheaper objective function. In addition, DAD is the best performing algorithm for real-world oriented experimental environments. Therefore, DAD is an appropriate algorithm to compare with the performance of our proposed algorithm \emph{Sequential Experimental Design via Reinforcement Learning} (RL-SED, in short). We also compare with the performance of a randomly chosen policy as a baseline.

In the following experiments, we consider three different problem settings. Each problem setting, i.e., each ``experimental environment" in the terminology of reinforcement learning, has been originally used by the previous works \cite{foster2021deep,huan2016sequential,Sheng2005MaximumLM}, respectively. In order to consider the realism of the experiment and the cost of the experiment, we have made several changes from the original experimental environments in this study. In the previous works, there were significant changes found in the consecutive designs, i.e., $a_t$ in the update \eqref{eq:design_diff} is quite large in magnitude, which leads to an unrealistic sequential design in practical settings. Even more, some did not consider the sequential nature of the experiments. 

In what follows, we describe the problem setting, focusing on the differences from the original experimental environments, and present the results for each of the three test cases, respectively. In every problem setting, we use CID \eqref{scid} as a quality criterion (reward) of experimental designs, and the number of episodes required for the learning until a sufficient convergent as a sample efficiency.

In each of the three experimental environments, the DAD and RL-SED agents use the state $s_{0:t-1}$ to determine the action $a_t$

Detailed information about the experimental environment and additional results should be written in the supplementary material.

\section{Location finding}

In this example, we construct an environment based on the experimental environment called ``Location finding" \cite{foster2021deep}. This is a problem to infer the locations of multiple unknown sound sources in space by sampling the intensity of the sound at a moving point. Our task here is to optimize a policy on the trajectory of the moving point. We assume an environment where three sound sources are randomly placed in location finding. The location of each source $\theta_i$  is sampled by $\theta \sim N(0,\sigma_1^2)$, where $N(0,\sigma_1^2)$ denotes a Gaussian distribution with mean 0 and variance $\sigma_1^2$.

In this environment, the agent outputs distance traveled from the current location as its action $a_t$. After updating the design $\xi_t$ according to the action, the agent observes the intensity of the sound. The sound intensity $\mu_t$ at the observation point $\xi_t$ is given by
\begin{align}
  \mu_t=b+\sum^{3}_{k=1}\frac{\alpha}{m+\parallel \theta_k - \xi_t \parallel^2}.
\end{align}
The sound observation contains noise. Therefore, the observed value $y_t$ can be obtained according to the following probability distribution.
\begin{align}
  y_t \sim N(\mu_t,\sigma_2^2)
\end{align}

In order to consider the cost of the experiment, we restrict the distance that the agent (i.e., the moving point) can travel during a time step to $d_1$. Furthermore, we also set a limit on the distance that the agent can travel in one episode to $d_2$. Thus, an episode ends when one of two conditions is met: either the max episode steps $T$ is reached, or the total travel distance exceeds $d_2$.

In order to consider the cost of the experiment, we restrict the distance that the agent (i.e., the moving point) can travel during one time step to $d_1$. This reduces the amount of change in the experimental designs between consecutive steps and makes the cost of the experiment not too large. We also set a limit on the distance that the agent can travel in one episode to $d_2$. Thus, an episode ends when one of two conditions is met: either the max episode steps $T$ is reached, or the total travel distance exceeds $d_2$. By putting these restrictions, we can consider two efficiencies simultaneously: the amount of the EIG from the experiment and the cost of experiment. These constraints can also be regarded as helpful in considering applications to real-world experimental environments.

In this experiment, we used Contrastive information bounds (CID) as the agent's reward, which is computed as follows:
\begin{align}
  \mathcal{I}_L=\log \frac{p(y_{0:t}|\theta_0,\xi_{0:t})}{\frac{1}{L+1} \sum^L_{l=0}p(y_{0:t}|\theta_l,\xi_{0:t})}.
\end{align}

The detailed parameters of the experimental environment are shown in Table \ref{tab:location}.

\begin{table}[th]
  \caption{Parameters in Location finding problem}
  \label{tab:location}
  \centering
  \begin{tabular}{rl}\toprule
     Parameter & Value\\ \midrule
     Standard deviation of sound source, $\sigma_1$ & 40\\
     $\alpha$ & 1\\
     Base signal, $b$ & 0.3 \\
     Max signal, $m$ & $10^{-4}$ \\
     Standard deviation of noise, $\sigma_2$ & 0.5 \\
     Max distance for each step, $d_1$ & 3 \\
     Max distance for each episode, $d_2$ & 50\\
     Max episode steps, $T$ & 100\\
     Inner sample size of CID, $L$ & 2000 \\ \bottomrule
  \end{tabular}
\end{table}

\begin{figure}[t]
  \centering
  \includegraphics[width=0.7\linewidth]{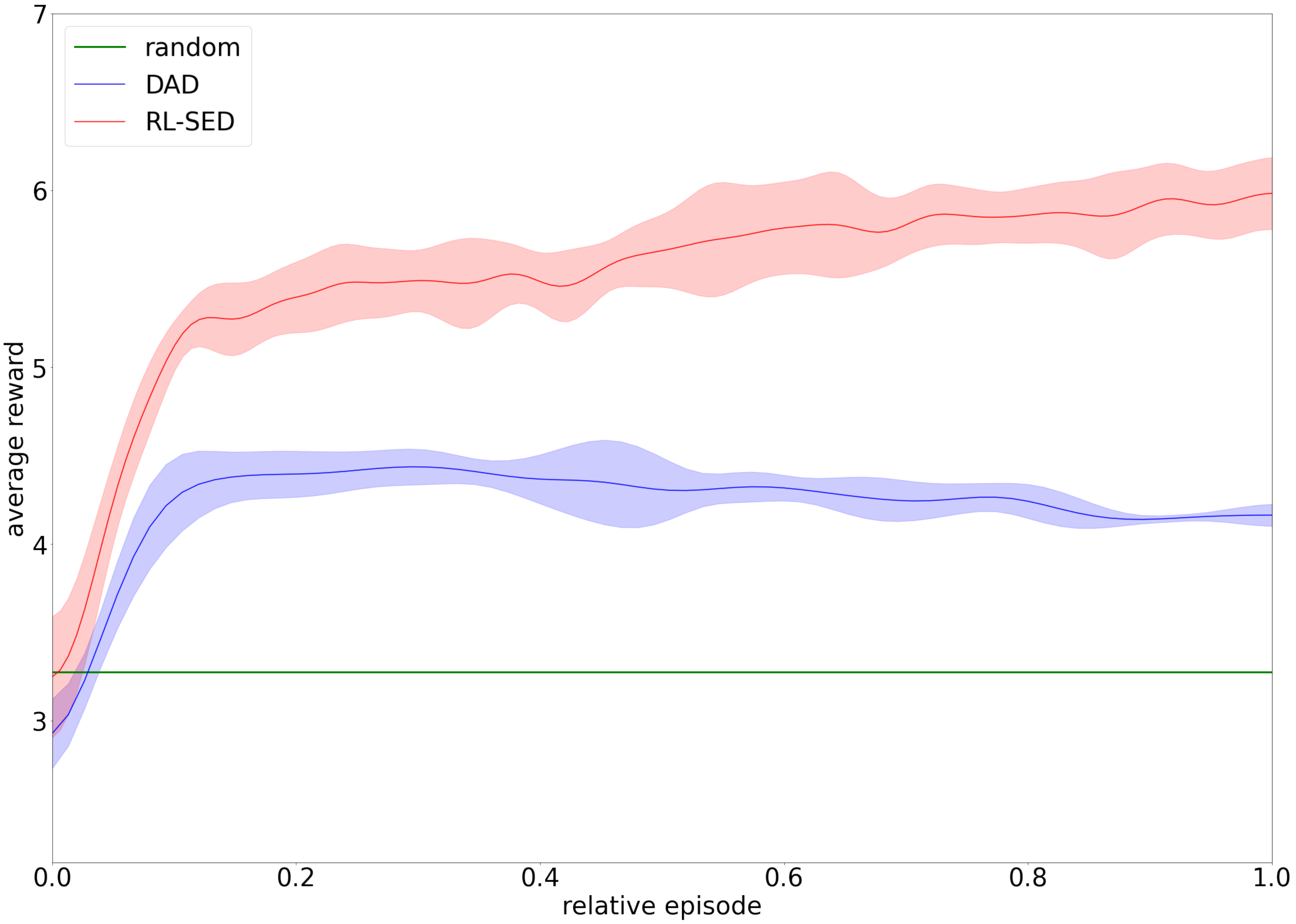}
  
  \caption{Learning curves for the location finding problem}
  \label{fig:location}
\end{figure}

Figure \ref{fig:location} shows the learning curves of random policy, DAD and RL-SED (ours). Table \ref{tab:locations} shows the average reward at the end of the episode and also the number of episodes required to achieve a sufficient convergence of the training. From Figure \ref{fig:location} and Table \ref{tab:locations}, we can see that our algorithm outperforms DAD on the average reward. In addition, our algorithm successfully completes the learning process within a quite fewer number of episodes. This is partially because our algorithm is based on off-policy learning, whereas DAD is based on on-policy learning.

\begin{table}[th]
  \centering
  \caption{Mean of final reward and total episode for training}
  \label{tab:locations}
  \begin{tabular}{rccc}\toprule
    environment & algorithm &reward & total episode\\ \midrule
    \multirow{3}{*}{Location finding} &random & 3.278  & - \\
    &DAD & 4.164  & 750000 \\
    & RL-SED & 5.955 & 15000\\ \midrule
    
    \multirow{3}{*}{Source inversion} &random & 3.586  & - \\
    &DAD & 5.051  & 40000 \\
    & RL-SED & 5.510 & 3000\\ \midrule
    
    \multirow{3}{*}{Death process} &random &  1.630  & - \\
    &DAD &  1.506  &1000000 \\
    & RL-SED & 2.042 & 15000\\ \bottomrule
  \end{tabular}
\end{table}

\begin{figure}[p]
  \centering
  \includegraphics[width=1.0\linewidth]{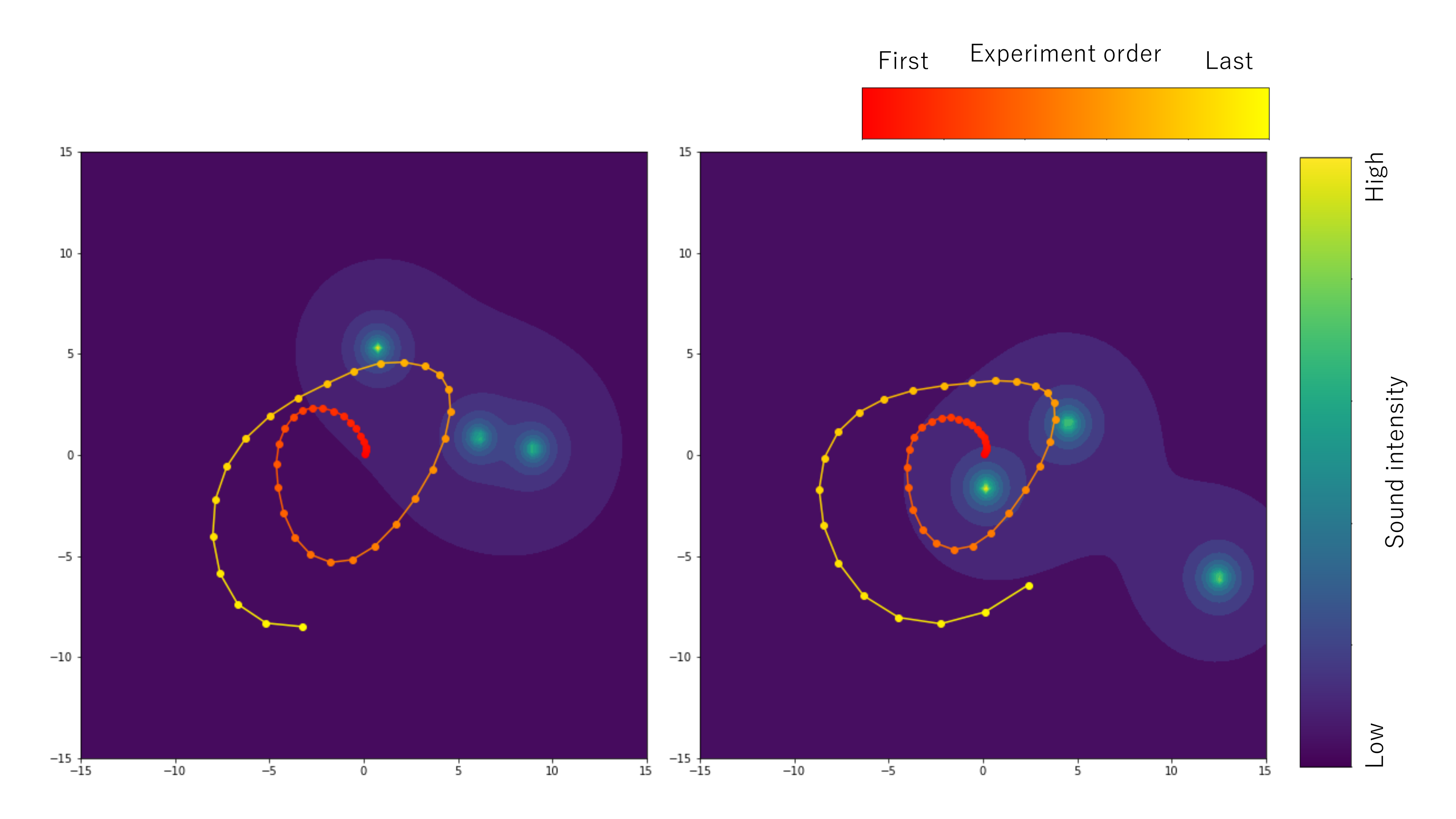}
  
  \caption{Sample trajectories of DAD}
  \label{trajectory_dad}
\end{figure}

\begin{figure}[p]
  \centering
  \includegraphics[width=1.0\linewidth]{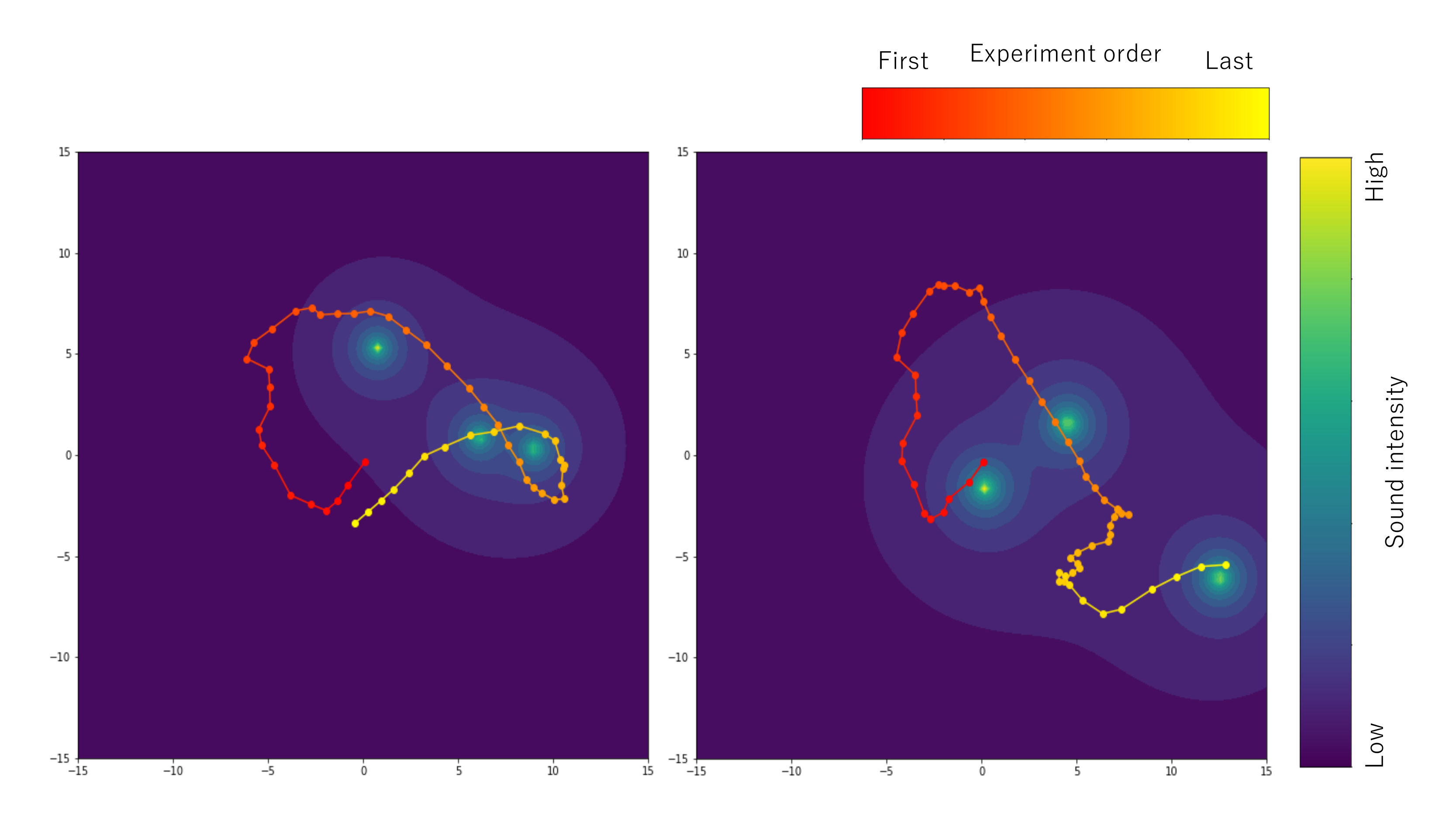}
  
  \caption{Sample trajectories of RL-SED}
  \label{trajectory_sac}
\end{figure}

Figures \ref{trajectory_dad} and \ref{trajectory_sac} show two examples of trajectories obtained by DAD and RL-SED, respectively. We can confirm from the figures how each algorithm samples in the two environments with different source locations.
As can be seen in Figure \ref{trajectory_dad}, DAD samples in a vortex-like fashion. In addition, the two trajectories have a similar shape, even though the positions of the sources are quite different. On the other hand, by comparing the two trajectories in Figure \ref{trajectory_sac}, we can see that both trajectories move in a similar way only at the beginning of the experiment. As the experiment progresses, we can see that they take different routes, as if they are trying to find the sound sources adaptively. This difference in the trajectories between DAD and RL-SED implies that RL-SED is able to capture the sequential nature of the experiment more accurately than DAD. As a result, RL-SED is able to adapt accurately even in situations where the locations of the sound sources change significantly, and to obtain higher rewards for identifying the locations of the sound sources.

We refer to this ability of the agent to respond to individual situations as the generalization performance. In order to quantitatively analyze the generalization performance, we consider different environments with different distributions $p(\theta)$ of the source locations, and conduct experiments. In the original environment for training agents, each of the three sources is independently sampled from a Gaussian distribution with mean 0 and standard deviation 40. In the additional experiments, we consider three environments with different standard deviations of 20, 40 (original), and 60 for the Gaussian distribution, and conduct experiments in each of them. The agents used in the experiments are the ones that achieved the highest performance during the training. 

\begin{table}[btp]
  \centering
  \caption{Result for different environments}
  \label{tab:different}
  \begin{tabular}{cccc}\toprule
     algorithm & environment  &reward & Ratio to the original\\ \midrule
    \multirow{3}{*}{random} &std=20 & 4.797  & 1.15 \\
                         &std=40 & 4.164  & 1.0 \\
                         &std=60 &  3.499  &0.840 \\ \midrule
    \multirow{3}{*}{DAD} &std=20 & 5.503  & 0.960 \\
                         &std=40 & 5.731  & 1.0 \\
                         &std=60 &  5.133  &0.900 \\ \midrule
    \multirow{3}{*}{RL-SED} &std=20 & 6.638 & 1.040 \\
                            &std=40 & 6.385 & 1.0 \\
                            &std=60 &   6.065 & 0.950\\ \bottomrule

  \end{tabular}
\end{table}

Table \ref{tab:different} shows the results of the additional experiments. As we can see, DAD has the best performance in the original environment, whereas RL-SED performs best when the standard deviation is 20 and the performance deteriorates as the standard deviation increases. As the results of the random policy show, in environments where the standard deviation is small, the distance between the sources is closer, which is expected to reduce the difficulty of the experiment. Due to the high generalization performance of RL-SED, it is able to achieve high performance even in environments different from those in which it was trained. On the other hand, the performance of DAD deteriorates in spite of the experiments in simple environments, which indicates a lower generalization performance of DAD. 
When the standard deviation is 60, the performance of each algorithm deteriorates, but the ratio of reduction from the original environment is smaller for RL-SED. These results suggest that RL-SED has a higher generalization performance than DAD.

From the quantitative results of the experiments as well as the visualization of the experimental designs, we can conclude that our algorithm works quite well in terms of both performance and sample efficiency and achieves a higher ability for sequential decision making than the existing methods.

\section{Source inversion}
We take an example called ``Contaminant source inversion problem" originally used in \cite{huan2016sequential}. In this experimental environment, we assume a situation in which chemical substances gradually diffuse in the air. During the experiment, the wind is blowing in space, and the chemicals move through space, diffusing as if swept by the wind. The location of chemical source $\theta$ and the strength of wind $w$ are sampled by following probability distributions.
\begin{align}
\theta \sim N(0,\sigma_1^2) \\
w \sim N(0,\sigma_2^2) \label{wind}
\end{align}

In this situation, the goal of the experimental design is to identify the source of the chemicals. Similarly to the location-finding problem, we consider measuring the concentration $y_t$ of the chemical at a moving point and our task is to optimize a policy on the trajectory of the moving point. The concentration of chemicals $\mu_t$ at observation point $\xi_t$ is given by
\begin{eqnarray}
    \mu_t &=&\frac{s}{\sqrt{2 \pi}(\sqrt{1.2+4Dt_d})}\exp \left( -\frac{\parallel \theta+d_w(t_d)-\xi_t \parallel^2}{2(1.2+4Dt_d)} \right) \\
    d_w(t_d)&=&10w(t_d-1)
\end{eqnarray}
where $t_d,d_w(t_d)$ are the total travel distance, and cumulative net displacement due to wind up to time step $t$, respectively. The total travel distance is the sum of the distances traveled by time $t$. In existing studies, the argument of $d_w$ is simply $t$. However, to consider the experiment's sequential nature, we have adopted the travel distance $t_d$ as the argument of $d_w$. Such a change implicitly imposes a cost on the agent for making a large move since this modification creates a situation where the material moves and diffuses more according to the amount of the move.

The observation contains noise. Therefore, the observed value $y_t$ can be obtained according to the following probability distribution.

\begin{align}
y_t \sim N(\mu_t,\sigma_3^2) 
\end{align}

In this source-inversion problem, the distance that the agent can travel during one observation step is restricted to $d_1$, as in location finding, so as to consider the cost of the experiment. We also set a limit on the distance that the agent can travel in one episode to $d_2$. In addition, although the direction and strength of the wind were fixed in the original experimental environment, they are generated randomly in our experimental environment as \eqref{wind}. Before this modification, since the direction and speed of the material moving are constant, it is possible to identify the source by moving the measurement point only in a specific direction. However, when both the strength and direction of the wind are random, the importance of sequential decision making increases. This is because if the wind strength varies, the concentration distribution of the material can be made broader as the material diffuses over time. Because of these characteristics, we consider the above modifications effective for more realistic sequential decision making problems.

In this experiment, we used Contrastive information bounds (CID) as the agent's reward.

The detailed parameters of the experimental environment are shown in Table \ref{tab:source}.

\begin{table}
  \caption{Parameters in Source inversion problem}
  \label{tab:source}
  \centering
  \begin{tabular}{rl}\toprule
     Parameter & Value\\ \midrule
     Standard deviation of chemical source, $\sigma_1$ & 10\\
     Standard deviation of wind, $\sigma_2$ & 0.1\\
     Concentration strength, s & 30\\
     Diffusion coefficient, D & 0.1\\
     Standard deviation of noise, $\sigma_3$ & 0.5 \\
     Max distance for each step, $d_1$ & 1 \\
     Max distance for each episode, $d_2$ & 50\\
     Max episode steps, $T$ & 100\\
     Inner sample size of CID, $L$ & 2000 \\ \bottomrule
  \end{tabular}
\end{table}

\begin{figure}
  \centering
  \includegraphics[width=0.7\linewidth]{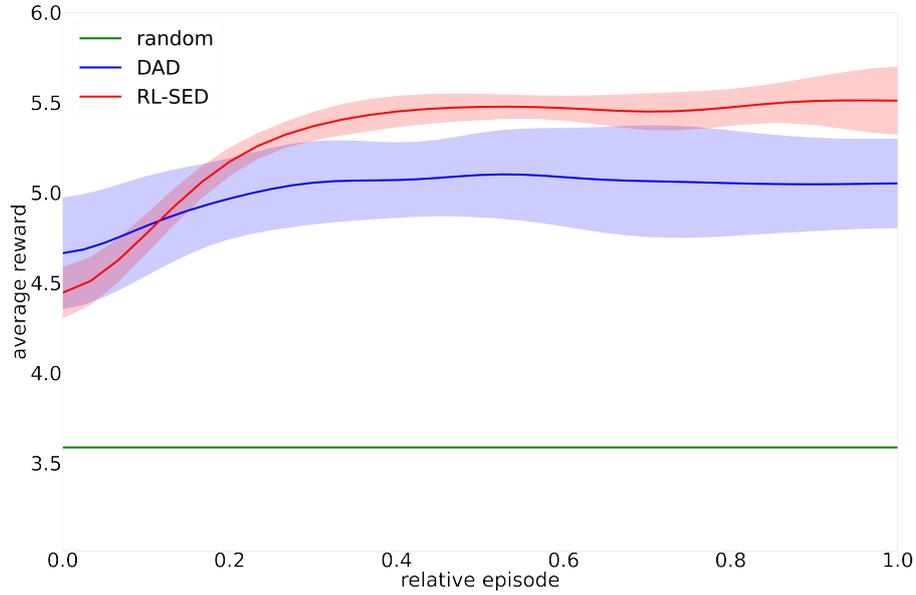}
  \caption{Learning curves for the source inversion problem}
  \label{fig:source}
\end{figure}

Figure \ref{fig:source} shows the learning curves of random policy, DAD and RL-SED (ours). Table \ref{tab:locations} shows the mean reward of each algorithm at the end of the training and the number of episodes used for training. From  Table~\ref{tab:locations}, we can see that our algorithm outperforms DAD on average. In addition, our algorithm requires fewer episodes for learning than DAD, indicating that the learning process is more efficient. 

We conducted additional experiments to evaluate the generalization performance of each model. In this experiment, we prepared three environments with different standard deviations of the Gaussian distribution to compare the agents' performance in environments with different distributions of the location of the chemicals. We use the agents who achieved the highest performance during the training. The results for each environment are shown in the table \ref{tab:diff_source}.

\begin{table}
  \caption{Result for different environments in source inversion problem}
  \label{tab:diff_source}
  \centering
  \begin{tabular}{cccc}\toprule
     algorithm & environment  &reward & Ratio to the original\\ \midrule
     \multirow{3}{*}{random} &$ \sigma_1$=5 & 4.716 & 1.31 \\
                         &$ \sigma_1$=10 & 3.586  & 1.0 \\
                         &$ \sigma_1$=15 &  3.465  &0.966\\ \midrule
    \multirow{3}{*}{DAD} &$ \sigma_1$=5 & 5.232 & 1.01 \\
                         &$ \sigma_1$=10 & 5.190  & 1.0 \\
                         &$ \sigma_1$=15 &  4.948  &0.953 \\ \midrule
    \multirow{3}{*}{RL-SED} &$ \sigma_1$=5 & 5.971 & 1.05 \\
                            &$ \sigma_1$=10 & 5.716 & 1.0 \\
                            &$ \sigma_1$=15 &   5.402 & 0.945\\ \bottomrule

  \end{tabular}
\end{table}

As the random policy results show, the experiment's difficulty increases as $\sigma_1$ increases.
The DAD and RL-SED agents performed best in the $\sigma_1=5$ environment and performed worst in the $\sigma_1=15$ environment. The increase in performance at $\sigma_1=5$ was higher for RL-SED, and the decrease in performance at $\sigma_1=15$ was smaller for DAD. However, the performance of RL-SED greatly exceeded that of DAD in all environments.

From the results and the properties of the experiments, we can claim that our algorithm outperforms the existing methods in terms of performance, sample efficiency, and the ability for sequential decision making.

\section{Death process}
The death process is an epidemiological model that describes the number of infected people \cite{Cook2008OptimalOT}, and the basic setup for this experiment is adapted from the previous works \cite{pmlr-v119-kleinegesse20a,foster2021deep}. In the death process example, the problem is to estimate the infection rate by observing the number of people infected with infectious diseases for which the probability of infection increases over time. Specifically, the infection probability $\eta$ is described by
\begin{align}
  \eta=1-\exp(-\xi\theta),
\end{align}
where $\xi$ denotes the time since the spread of infection and $\theta$ is a random latent variable representing the infectivity obtained by $\theta \sim \mathrm{TruncatedNormal}(\mu_{\theta},\sigma_{\theta},\min = 0 ,\max = \infty)$. In the design of the experiment, the interval time $a_t$ from the time of the previous observation is the output of the action in \eqref{eq:design_diff}. In the first observation, the output $a_0$ is used as the first observation time $\xi_0$. In every observation, a certain number of people are inspected. The number of infected people in the test is given as the observed value $y_t$, and the following distribution determines $y_t$.
 \begin{align}
     y_t \sim \mathrm{Binominal}(N,\eta)\\
     \eta=1-\exp(-\theta \xi_t )   
 \end{align}

In the death process, to make the experimental environment more realistic, $\xi_t$ should be monotonically increasing as a function of $t$, whereas, in the existing works, $\xi_t$ can be freely chosen to set up a sequential problem. However, in our experimental environment, by putting the constraint $a_t>0$, $\xi_t$ is forced to be monotonically increasing. To set these constraints, we used the Softplus function for the last layer of each agent.

In this experiment, we used CID as the agent's reward.

The detailed parameters of the experimental environment are shown in Table \ref{tab:death}.

Table \ref{tab:locations} shows the mean reward of the performance of each algorithm at the end of the training and the number of episodes required for training. From Table \ref{tab:locations}, we can see that our algorithm again outperforms DAD on the mean reward. The table also shows that DAD has a lower performance than random policy, indicating a failure in learning a good policy in this example. 

\begin{table}
  \caption{Parameters in death process problem}
  \label{tab:death}
  \centering
  \begin{tabular}{rl}\toprule
     Parameter & Value\\ \midrule
     Mean of infection rate, $\mu_{\theta}$ & 1.0\\
     Standard deviation of infection rate, $\sigma_{\theta}$ & 1.0\\
     Number of observations, $T$ & 4 \\
     Number of infected people, $N$ & 50\\
     Inner sample size of CID, $L$ & 2000 \\ \bottomrule
  \end{tabular}
\end{table}

\begin{figure}
  \centering
  \includegraphics[width=0.8\linewidth]{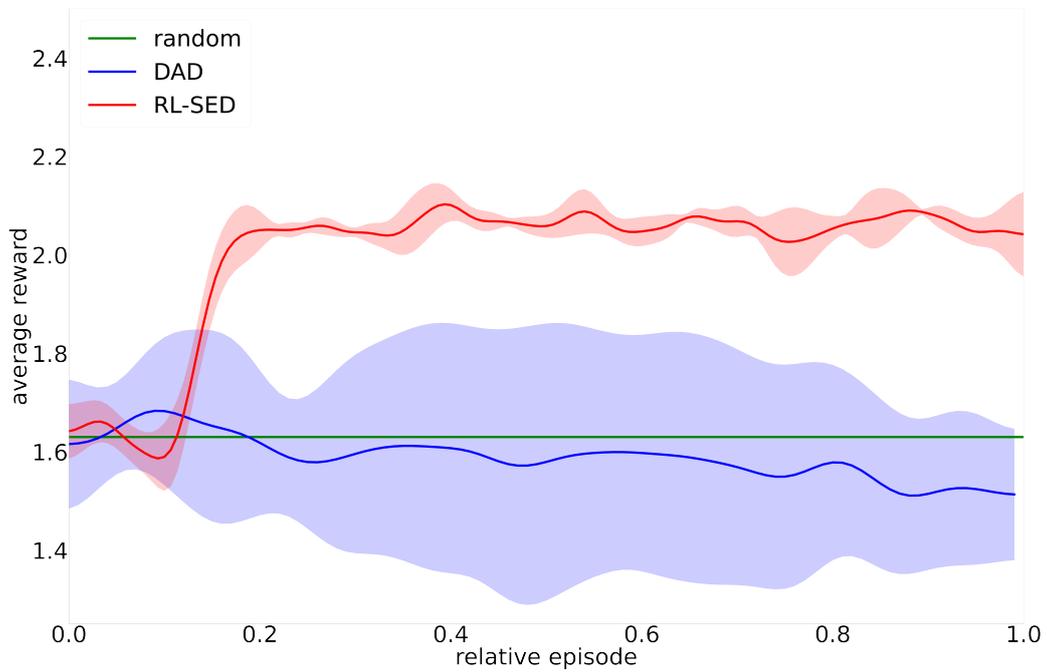}
  \caption{Learning curves for the death process problem}
  \label{fig:death}
\end{figure}

We conducted additional experiments to evaluate the generalization performance of each model. In this experiment, we prepared three environments with different means of the distribution to compare the agents' performance in environments with different distributions of the infection rate. We use the agents who achieved the highest performance during the training. The results for each environment are shown in Table \ref{tab:diff_death}.

\begin{table}
  \caption{Result for different environments in death process problem}
  \label{tab:diff_death}
  \centering
  \begin{tabular}{cccc}\toprule
     algorithm & environment  &reward & Ratio to the original\\ \midrule
     \multirow{3}{*}{random} &$ \mu_{\theta}$=0.5 & 1.553 & 0.953 \\
                         &$ \mu_{\theta}$=1.0 & 1.630  & 1.0 \\
                         &$ \mu_{\theta}$=1.5 &    1.680  &1.03 \\ \midrule
    \multirow{3}{*}{DAD} &$ \mu_{\theta}$=0.5 & 1.553 & 0.989 \\
                         &$ \mu_{\theta}$=1.0 & 1.570  & 1.0 \\
                         &$ \mu_{\theta}$=1.5 &  1.391  &0.886 \\ \midrule
    \multirow{3}{*}{RL-SED} &$ \mu_{\theta}$=0.5 & 2.256 & 1.04 \\
                            &$ \mu_{\theta}$=1.0 & 2.163 & 1.0 \\
                            &$ \mu_{\theta}$=1.5 &   1.965 & 0.908\\ \bottomrule

  \end{tabular}
\end{table}

Figure \ref{fig:death} shows the learning curves of random policy, DAD and RL-SED (ours). As shown in Table \ref{tab:diff_death}, the performance of the random policy improved as the mean value $\mu_{\theta}$ increased. In contrast, the performance of DAD and RL-SED improved as $\mu_{\theta}$ became smaller. These results suggest that it is challenging to explain the difficulty of the experiment by the size of $\mu_{\theta}$. However, the difficulty in explaining the difficulty of the experiment is not a problem in evaluating the generalization performance of the agent.

The DAD performed best at $\mu_{\theta}=1.0$ where it was trained. However, at $\mu_{\theta}=0.5$ and $\mu_{\theta}=1.5$, the performance of the agent decreased. The performance of DAD was the same as that of random policy at $\mu_{\theta}=0.5$, and lower than that of random policy in all other environments. From this, we can argue that DAD has poor generalization performance on death process problems and that it does not learn well. RL-SED achieved the best performance at $\mu_{\theta}=0.5$. In addition, at $\mu_{\theta}=1.5$, the performance decreased compared to the trained environment, but the reduction rate was lower than that of DAD.

From the above results, we can claim that RL-SED can learn better than existing algorithms in the death process problem and has higher generalization performance.

\chapter{Conclusion}
In this paper, we have proposed a novel method \emph{Sequential Experimental Designs via Reinforcement Learning} to efficiently learn a policy to construct Bayesian experimental designs in a sequential manner. With some additional modifications to the previously-studied examples, we have enabled to quantify the cost of the experiment and sample efficiency, both of which are essential aspects in applying the experimental designs to real-world experiments. Our proposed method successfully learns a policy with higher rewards than the compared methods in all three experimental environments tested. Moreover, we showed that our approach is more stable and the sample efficiency is also better. In conclusion, this present work can make a significant contribution to the application of experimental designs to real-world, large-scale, complex problems.

As we have discussed, the concepts of experimental design and reinforcement learning have very similar characteristics in terms of the interaction between the agent and the environment. In this paper, we have connected these two (apparently different) domains and confirmed the effectiveness of our method. It might be interesting to transfer more state-of-the-art techniques in reinforcement learning to construction of sequential experimental designs.

Besides, although the proposed method performed quite well in many aspects as compared to the known methods, the use of $s_{0:t}$ for sequential decision making resulted in huge state space. The size of the state space is a problem that must be solved when assuming a more complex experimental environment. It will be necessary to investigate how to reduce the state space given to the policy in the future.

\bibliographystyle{jplain}
\bibliography{library}


\end{document}